\documentclass[runningheads]{llncs}
\usepackage[T1]{fontenc}
\usepackage{graphicx}
\usepackage{svg}
\usepackage{amsmath}
\usepackage{amssymb}
\usepackage{array}
\usepackage{geometry}
\usepackage{booktabs} % for professional tables
\usepackage{hyperref}

\begin{document}
\title{Failures Are the Stepping Stones to Success: Enhancing Few-Shot In-Context Learning by Leveraging Negative Samples}
% \author{
%     Yunhao Liang\inst{1,2}\orcidID{0009-0002-8104-0903} \and
% Xinran Xie\inst{1,2}\orcidID{0009-0009-9546-5022} \and
% Mengke Chen\inst{1,2}\orcidID{0009-0004-9244-0982} \and
% Zhe Cui\inst{1,2}\orcidID{0000-0002-9708-2000}
% }
%
\author{
  Yunhao Liang\inst{1,2}\orcidID{0009-0002-8104-0903} \and
  Ruixuan Ying\inst{3}\orcidID{0009-0008-9597-599X} \and
  Takuya Taniguchi\inst{3}\orcidID{0000-0001-9865-1351} \and
  Zhe Cui\inst{1,2}\orcidID{0000-0002-9708-2000}
}
\institute{Chengdu Institute of Computer Applications, Chinese Academy of Sciences, Chengdu 610041, China \and
University of Chinese Academy of Sciences, Beijing 101408, China \and
Institute of Multidisciplinary Research for Advanced Materials (IMRAM), Tohoku University Sendai 980-8577, Japan\\
\email{{liangyunhao22@mails.ucas.ac.cn, ying.ruixuan.s5@dc.tohoku.ac.jp}}
}
% \url{http://www.springer.com/gp/computer-science/lncs}
\maketitle              % typeset the header of the contribution
\begin{abstract}
    Large Language Models exhibit powerful few-shot in-context learning (ICL) capabilities, but the performance is highly sensitive to provided examples.
    Recent research has focused on retrieving corresponding examples for each input query, not only enhancing the efficiency and scalability of the learning process but also mitigating inherent biases in manual example selection.
    However, these studies have primarily emphasized leveraging Positive samples while overlooking the additional information within Negative samples for contextual learning. 
    We propose a novel method that utilizes Negative samples to better select Positive sample examples, thereby enhancing the performance of few-shot ICL. Initially, we construct Positive and Negative sample corpora based on Zero-Shot-Cot. Then, during inference, we employ a semantic similarity-based approach to select the most similar examples from both the Positive and Negative corpora for a given  query. Subsequently, we further retrieve Positive examples from the Positive sample corpus based on semantic similarity to the Negative examples, then  concatenating them with the previously selected Positive examples to serve as  ICL demonstrations. Experimental results demonstrate that our approach surpasses methods solely relying on the most similar positive examples for context, validating that the additional information in negative samples aids in  enhancing ICL performance through improved Positive sample selection.

\keywords{Prompt Engineering \and Negative Samples \and In-Context-Learning.}
\end{abstract}
\section{Introduction}

Large language models (LLMs) are characterized by their massive parameter sizes and extensive training data,  which enable them to encapsulate rich and profound knowledge \cite{touvron2023llama,radford2019language,zhao2023survey}.
They have exhibited significant performance across diverse applications and reasoning tasks such as question answering, code generation, and mathematical reasoning \cite{yang2024harnessing,liang2025exploring,liang2025recode}. 
Prompting  learning methods \cite{reynolds2021prompt} such as in-context learning \cite{radford2019language}  and chain-of-thought \cite{wang2022self,chen2022program} have been introduced to empower LLM to undertake few-shot or even zero-shot learning, facilitating adaptation to new scenarios with minimal or absent annotated data.
Few-shot In-Context Learning (ICL), initially introduced alongside the emergence  of GPT-3, stands as a novel paradigm in the era of LLMs. It denotes the model's capability to assimilate insights from a few contextual exemplars: furnishing the model with specific instances (referred to as demonstrations.) of a particular task alongside the questions to be examined, the model can furnish answers  to the test questions grounded on the provided examples. 
LLMs can  execute various intricate reasoning tasks through ICL. 
Furthermore, the entire process circumvents the necessity of updating model parameters; solely forward inference is conducted, resulting in a substantial reduction in the overhead associated with adapting large models to novel tasks. 
However, the efficacy of ICL is markedly sensitive to the configuration of examples, encompassing prompt templates, the curation of contextual examples, and the sequencing of examples \cite{ye2022unreliability}. 
Exploring methodologies to more effectively harness, the wealth of knowledge ingrained in LLMs through pre-training via ICL represents a burgeoning area of research.

Prior studies have explored a number of different techniques for selecting examples to construct few-shot ICL demonstrations based on similarity retrieval, as well as selecting the most complex or representative samples from the training set for inference ~\cite{luo2023dr,rubin2021learning,lu2021fantastically}.
However, these studies primarily focused on similarity metrics or retrieval methods to identify the best examples for prompt demonstrations, all of which were positive examples from the corpus.
To the best of our knowledge, no study has investigated how to leverage additional valuable information from negative examples to aid in constructing few-shot prompt demonstrations.
Negative samples, while often perceived as worthless as they fail to yield the correct answer, actually harbor valuable information, as addressed later.

Drawing inspiration from the human learning process, as we progress in our academic journey, we often confront similar exercises, some of which we answer correctly while others lead to errors. 
By focusing on rectifying errors in exercises, as cataloged in our error logs, we can enhance our learning process. In instances where we consistently falter in a particular problem category, additional practice with similar problems proves advantageous. Thus, we propose a method for constructing few-shot ICL demonstrations using negative samples. 
Our proposed method comprises two stages: corpus construction and demonstration construction for test queries. 
To ensure the retrieval of sufficiently similar examples for each test query, we initially cluster the dataset and randomly select half of the data from each cluster as the initial corpus. 
Subsequently, the model performs inference on these data, and based on whether the predicted results match the true labels, we categorize them into positive and negative samples and add them to the positive and negative corpora, respectively. 
The second stage involves constructing demonstrations for test query questions, also consisting of two steps. 
Firstly, we retrieve similar positive and negative examples from the positive and negative corpora for the test query, respectively.
Then, we retrieve new similar positive examples for negative samples from the positive corpus. Combining these positive examples acquired from both two retrieval steps yields our final demonstration. 
We believe that the new positive examples retrieved based on negative samples can be viewed as implicit error correction data, eliminating the need for explicit utilization of stronger LLM like GPT-4 for error correction \cite{an2310learning}. 
Experimental results across three inference tasks on seven datasets demonstrate the efficacy of our method in utilizing negative samples to enhance few-shot ICL performance.
To summarize, our work makes the following contributions:
\begin{itemize}
    \item We introduce a two-stage framework that utilizes negative samples to enhance few-shot ICL performance.
    \item Experimental results across multiple inference tasks and datasets demonstrate the efficacy of the proposed method. Furthermore, the results also prove that in-context learning can benefit from negative samples.
\end{itemize}

\section{Method}
Our proposed method consists of two stages. The first stage involves the construction of the corpus. 
The second stage leverages the corpus and utilizes the retrieved nega-tive sample instances to assist in the selection of positive sample instances during the construction of prompt demonstrations.

\subsection{Retrieval corpora construction}
\begin{figure}[!htbp]
    \centering
    \includegraphics[width=1.0\textwidth]{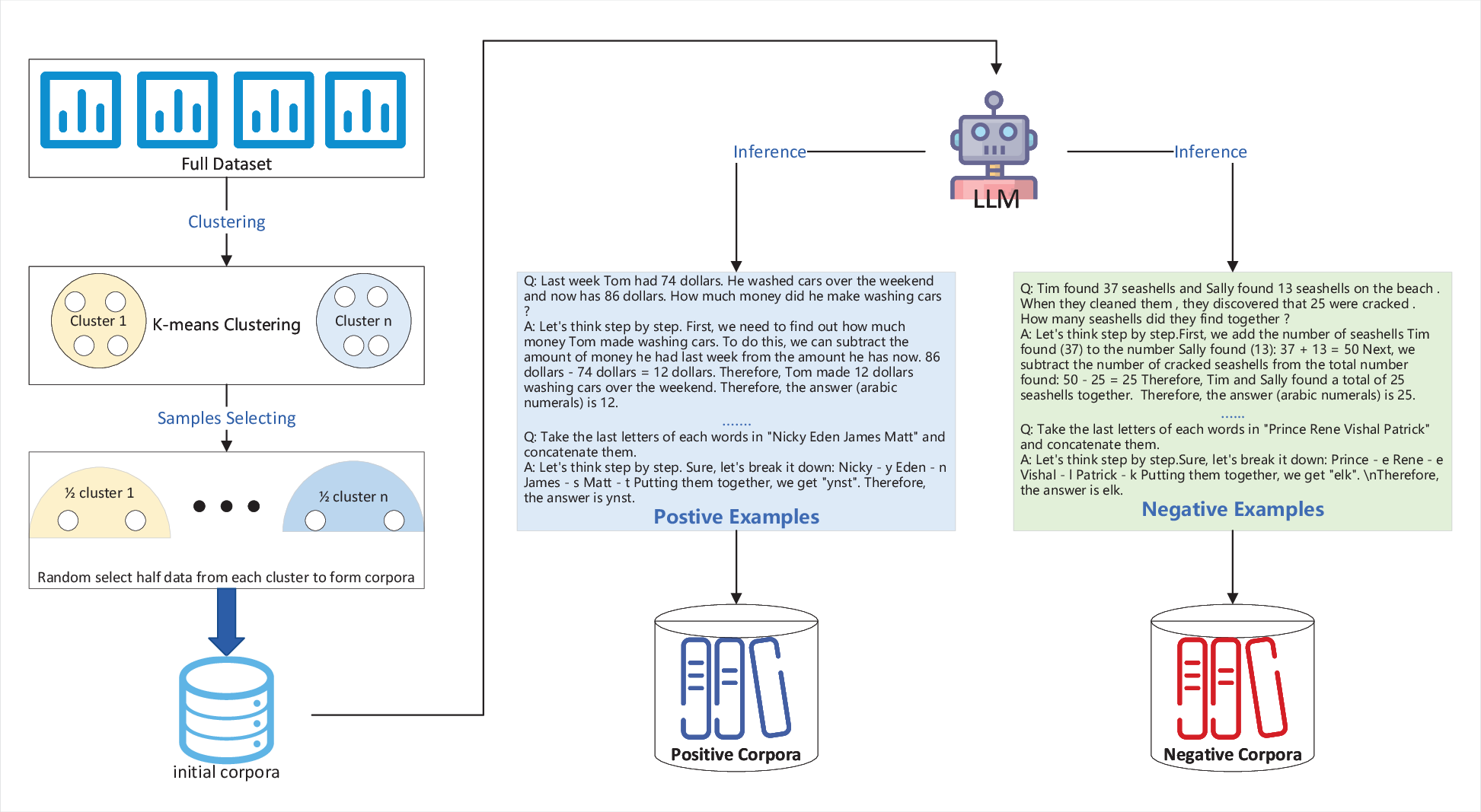}
    \caption{The process of corpus construction.}
    \label{fig1}
\end{figure}

In order to guarantee the availability of sufficiently similar instances for each test query question, the process of constructing the retrieval corpus began by conducting k-means clustering on the entire dataset. 
Subsequently, half of the samples from each cluster were randomly allocated to the training set $D_train$, while the other half formed the test set $D_test$. 
To create the positive and negative corpora for semantic similarity retrieval, Zero-Shot-CoT \cite{kojima2022large} is utilized to facilitate LLM in conducting inference on the training set. 
Samples were classified into positive or negative categories based on the agreement between their predicted results and the gold answer, and then appended to the corresponding positive and negative corpora. 
The process of constructing the corpus is illustrated in figure \ref{fig1}.

\subsection{Negative samples enhance demonstrations construction}
\begin{figure}[!htbp]
    \centering
    \includegraphics[width=1.0\textwidth]{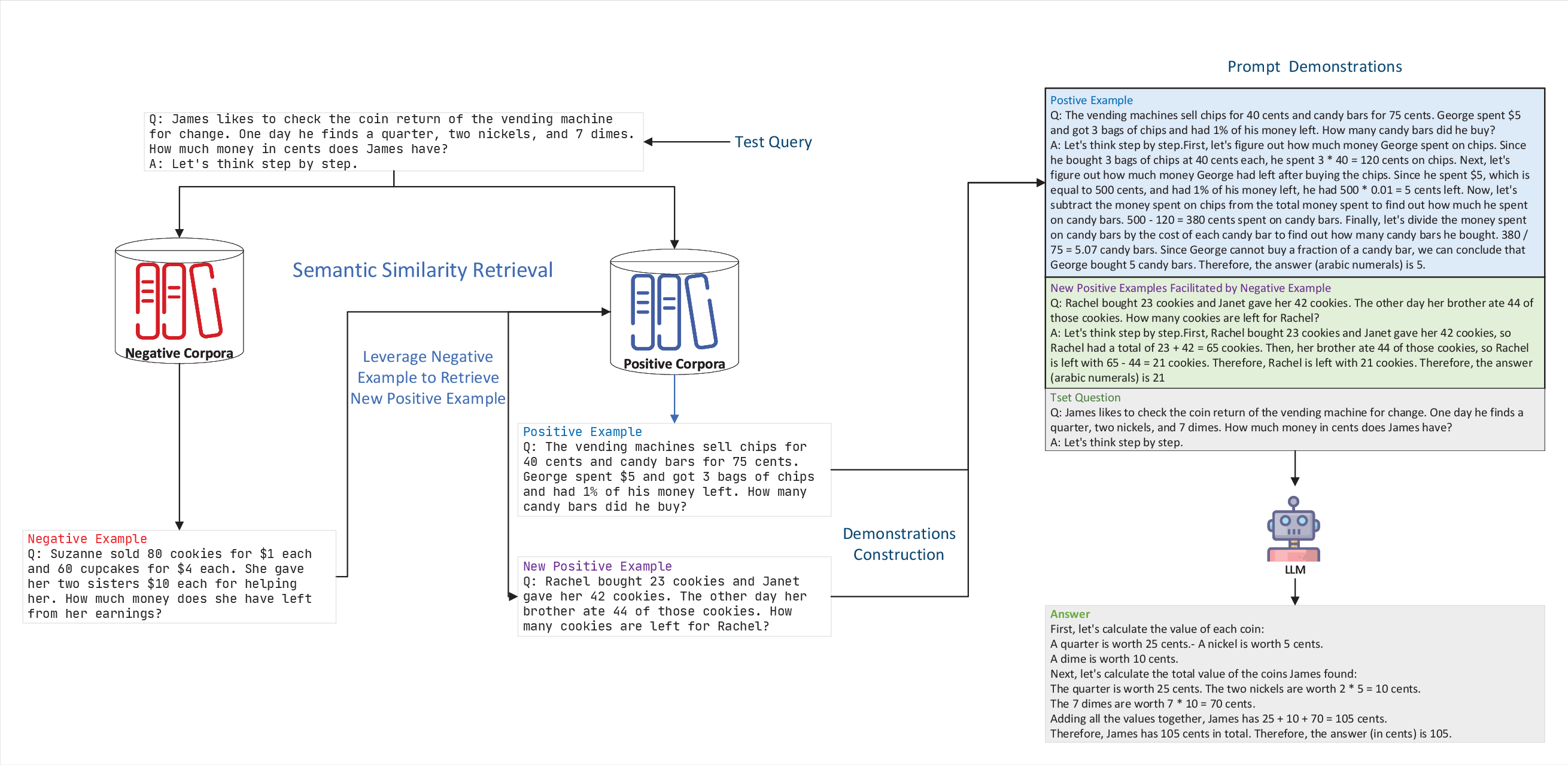}
    \caption{The process of constructing demonstrations with the aid of negative samples.}
    \label{fig2}
\end{figure}

Figure \ref{fig2} illustrates the procedure of utilizing negative samples to support the construction of few-shot ICL demonstrations, leveraging the established positive and negative corpora. 
For the given input query $q$, the final demonstration we present contains $k$ examples. 
Our approach to constructing demonstrations with the aid of negative samples encompasses the subsequent two steps. 
To begin with, we retrieve $k/2$ positive examples $Example_{pos}$ and $k/2$ negative examples $Example_{neg}$ from respective positive and negative corpora based on semantic similarity.
Subsequently, we utilize negative samples to assist in the selection of positive examples.
Specifically, for each retrieved negative example $Example_{(neg_i )} (i=0,1…k/2)$, we further search the positive corpus to find the most semantically similar positive sample $Example_{(pos-new_i )} (i=0,1…k/2)$. 
Ultimately, we concatenate these positive samples obtained through negative sample retrieval with the positive samples retrieved based on the query q to create the final demonstrations for the few-shot In-Context-Learning prompt.

\section{Experiments And Results}
\subsection{Experimental Setup}
\subsubsection{Tasks and Datasets.} 
Our proposed approach was assessed across seven benchmark datasets, corresponding to three different reasoning tasks: arithmetic reasoning, common sense reasoning, and symbolic reasoning. 
The datasets AddSub \cite{hosseini2014learning}, GSM8K \cite{cobbe2021training}, SingleEq \cite{koncel2015parsing}, and SVAMP \cite{patel2021nlp} were utilized for the arithmetic reasoning tasks. 
The tasks of commonsense reasoning employed the CommonsenseQA \cite{talmor2018commonsenseqa} and StrategyQA \cite{geva2021did} datasets. 
The Last Letter Concatenation \cite{wang2022self} dataset was the only one used for symbolic reasoning tasks, refraining from using the Coin-Flip due to its oversimplification and inability to effectively evaluate our method.

\subsubsection{Implementation Details.} 
We utilized GPT-3.5-Turbo  as the LLM for evaluating our proposed approach and comparison methods, which was accessed through OpenAI's paid API. Due to OpenAI's pricing policy, the number of prompt demonstrations k was simply set to 2. 
For each test question, we initially retrieved $\frac{k}{2}$ examples from both positive and negative sample corpora. 
Subsequently, for each negative sample example, we retrieved the most semantically similar positive sample example from the positive sample corpus. 
These two sets of positive sample examples were concatenated to form the demonstration for few-shot ICL. 
The number of clusters used during corpus construction varied depending on the number of samples included in each dataset: AddSub (4), SingleEq (4), Last Letter Concatenation (4), GSM8K (8), CommonsenseQA (8), SVAMP (6), and StrategyQA (10). 
For the sentence embedding model, we use Sentence-BERT \cite{reimers2019sentence} for encoding both query questions and candidate sentences contained within the corpus.

\subsubsection{Baselines.} 
We compared our proposed method against the following baseline methods: Few-Shot-CoT, Similarty-Based Few-Shot CoT, Contrastive Few-Shot CoT, and Random Few-Shot-CoT. For Few-Shot-CoT, we followed the setup of previous work \cite{kojima2022large}, 
where we simply appended "Let's think step by step." after the question as a CoT trigger and used their proposed answer cleaning technique to extract the final predicted output answer. 
For Similarty-Based Few-Shot CoT, we retrieved k semantically similar examples from the positive sample corpus for each test question and concatenated them to serve as prompt demonstrations. 
For Contrastive Few-Shot CoT, we retrieved $k/2$ most similar samples from both the positive and negative sample corpora and inserted them into the instruction template "Below is a positive example and a negative example. 
Learn from the positive example and avoid making the mistakes in the negative example. 
Here is the positive example: [positive example], And here is the negative example: [negative example]" as prompt demonstrations. 
For Random Few-Shot-CoT, we randomly selected two examples from the positive sample corpus to serve as demonstrations.

\subsection{Results}
\subsubsection{Performance of our proposed method.}
Table \ref{tab1} illustrates that our proposed approach, which utilizes negative samples to select better positive examples for prompt instances, consistently achieves superior performance compared to all baseline methods across all seven datasets, encompass-ing three distinct reasoning tasks. 
Our method consists of two variants. The first, (1 pos+1 pos-new), involves retrieving a positive example and a negative example from the positive and negative sample corpora, respectively. 
Based on the negative exam-ple, we then retrieve the most similar new positive example from the positive sample corpus. 
These two positive examples serve as the instances. The second variant, (2 pos-new), involves initially retrieving the two most similar negative sample instances to the current test question from the negative sample corpus. 
We then retrieve the most similar positive sample instance for each of these two negative sample instanc-es from the positive sample corpus. These two new positive sample instances serve as the demonstration.

\begin{table}[!htbp]
    \centering
    \caption{Accuracy on seven datasets from three categories of reasoning tasks.}\label{tab1}
    \begin{tabular}{|l|l|l|l|l|l|l|l|}
    \hline
        Method & \multicolumn{6}{c}{Dataset} ~ &\\ \hline
        ~ & AddSub & GSM8K & SingleEq & SVAMP & CommonSenseQA & StrategyQA & Last Letter \\ \hline
        Zero-Shot-CoT & 85.5 & 78.3 & 93.7 & 81.8  & 73.6 & 66.1 & 41.8  \\ 
        Similarity Few-Shot & 83.0 & 73.1 & 92.1 & 81.8  & 75.6 & 63.1 & 67.7  \\ 
        Contrastive CoT & 83.0 & 71.7 & 92.1 & 77.6  & 71.5 & 61.8 & 67.7  \\ 
        Random Few-Shot & 85.5 & 77.7 & 94.1 & 81.8  & 75.7 & 65.8 & 76.0  \\ 
        Ours(1 pos-new) & {\bfseries 87.6} & {\bfseries 78.6} & {\bfseries 94.8} & 83.7  & 75.2 & 66.0 & 76.5  \\ 
        Ours(2 pos-new) & 83.6 & 77.4 & 94.1 & {\bfseries 84.1}  & {\bfseries 75.9} & {\bfseries 67.2} & {\bfseries 83.3}  \\ \hline
    \end{tabular}
\end{table}

\subsubsection{Optimal selection of negative sample quantities and their applicability across various tasks.} 
To discern the best count of negative samples to be used in our prompting approach, we undertook a quantitative analysis.
The goal was to determine the number of most similar negative samples that should be incorporated into the prompt demonstration to assist in locating the most similar positive examples. 
We selected a representative dataset for each of the three reasoning tasks to conduct experiments. 
For mathematical reasoning problems, we used AddSub. For common sense reasoning, we chose CommonsenseQA. 
As for symbolic reasoning, we used Last Letter Concatenation. We denote the total number of demonstration examples as total, the count of positive sample examples as m, and the count of negative sample examples (i.e., new positive sample examples retrieved based on negative samples) as n. 
The sum of the positive sample examples and the negative sample examples equals the total number of demonstration examples (i.e. m+n=total). 
We chose a total of 6 examples, conducting experiments for combinations ranging from (m=0,n=6) to (m=6,n=0). 
The results of these experiments are presented in Table 2. 
From the results, we can observe that our method, which combines the most similar positive examples and the positive examples most semantically like the negative examples, improves accuracy compared to using only the most similar positive examples or only the positive examples most similar to the negative examples. 
However, using too many negative examples can lead to a performance decline (as seen in Last Letter when using 4 and 6 negative examples, which caused a decrease in accuracy). 
We attribute this to semantic drift, where additional negatives may be only weakly related to the query, pulling the second-stage retrieval toward less relevant positives.
Interestingly, we observe that employing only newly retrieved positive examples via negative sample retrieval yields the most promising performance in CommonSenseQA.
This corroborates our earlier observation that demonstrations composed exclusively of positives retrieved via negative examples outperform mixtures of traditional and negatively-anchored positives, suggesting that commonsense reasoning benefits more from dense, error-driven corrective signals than from balanced coverage.
The analysis of the table suggests that for arithmetic and symbolic reasoning tasks, the optimal number of negative samples to use is two, whereas for commonsense reasoning problems, employing exclusively newly retrieved positive examples through negative sample retrieval results in the highest performance. 
We hypothesize that this discrepancy stems from the nature of the required knowledge: arithmetic and symbolic tasks demand a small, explicit correction of procedural steps, so a limited injection of error-driven positives prevents over-correction; 
in contrast, the open-ended, associative nature of commonsense questions benefits from a denser set of error-driven positives that collectively tighten the decision boundary around less sharply defined concepts.
Furthermore, our approach demonstrates excellent applicability across the three reasoning tasks, as evidenced by substantial performance improvements observed across these three representative tasks.

\begin{table}[!htbp]
    \centering
    \caption{The proper number of negative examples and suitable reasoning task.}\label{tab2}
    \begin{tabular}{|l|l|l|l|}
    \hline
        Num of N and M & AddSub & Last Letter & ComonSenseQA  \\ \hline
        m=0, n=6  & 85.1 & 79.6 & {\bfseries 80.0}  \\ 
        m=1, n=5 & 85.1 & 82.4 & 75.2  \\ 
        m=2, n=4 & 87.5 & 79.2 & 77.0  \\ 
        m=3, n=3 & 86.5 & 82.1 & 77.2  \\ 
        m=4, n=2  & 88.6 & {\bfseries 83.7} & 75.8  \\ 
        m=5, n=1 & {\bfseries 90.1} & 83.3 & 76.8  \\ 
        m=6, n=0 & 84.0 & 82.1 & 78.3  \\ \hline
    \end{tabular}
\end{table}

\section{Related Work}
\subsubsection{Retrieve similar examples as demonstrations.} 
Rather than employing a static prompt, recent work has shifted to dynamic retrieval of task-specific exemplars at inference time. 
\cite{agrawal2022context} focus on machine-translation tasks and show that jointly optimizing lexical and semantic similarity yields demonstrably better translations than either metric alone. 
\cite{ye2023compositional} go beyond single-sentence retrieval and propose compositional exemplars that are assembled from multiple, smaller snippets so that the final demonstration covers the full structural pattern of the query; 
this is the first study to treat demonstration construction as a combinatorial search problem. 
\cite{mishra2022towards} introduce a continual memory bank that stores user feedback and incrementally re-indexes retrieved examples, enabling the retriever itself to evolve with new error signals. 
\cite{li2023mot} present Memory-of-Thoughts (MoT), which first “pre-thinks” a set of candidate reasoning chains, then recalls the most relevant chains via a lightweight non-parametric memory, effectively decoupling generation and retrieval of rationales.
Despite these advances, these work exclusively retrieve positive (correct) examples and optimize similarity metrics or memory structures. 
Our method is the first to use negative (incorrect) examples as semantic anchors to mine additional positive demonstrations, thereby shifting the focus from “finding the closest correct pattern” to “correcting the closest incorrect pattern.”
% Several previous studies have already demonstrated that retrieving samples most similar to the current test query as examples can enhance the ability of ICL. ~\cite{agrawal2022context,ye2023compositional,mishra2022towards,li2023mot}
% Moreover, these works delved into how performance is affected by the utilization of diverse similarity metrics and retrieval techniques.

\subsubsection{Learning from negative samples.}
According to \cite{an2023learning,an2024can}, there is a proposition regarding whether the reasoning prowess of LLMs can be augmented through a reverse learning approach, wherein learning occurs from the model's errors. 
The concept of learning from errors is remi-niscent of the learning methodologies observed in human students. 
Upon encounter-ing difficulties in problem-solving, individuals engage in error analysis to understand their mistakes and rectify them. 
Implementing this paradigm of learning from past mistakes holds promise for enhancing the model's reasoning prowess.

Recent investigations have uncovered that models fine-tuned on positive and negative samples separately show a limited overlap in accurately addressed cases \cite{li2024turning}. 
This observation highlights the presence of crucial additional information within neg-ative samples, which is lacking in positive samples. 
They introduce a technique aimed at harnessing negative sample data to extract intricate reasoning capabilities, thereby enhancing the inference prowess of language models through Negative Assistant Training (NAT) and Negative Calibrated Enhancement (NCE). 
Furthermore, they present the Adaptive Self-Consistency (ASC) method, leveraging negative samples to refine the model's inference mechanism.

\section{Conclusion}
LLMs have exhibited exceptional natural language understanding and inference capabilities, with context learning being one of their prominent abilities. Previous research has predominantly depended on fixed, manually curated examples or random examples. More recent research has shifted focus towards using examples similar to the test question for few-shot demonstrations, yet these methods often neglect the potential valuable information embedded in negative samples.
In this work, we investigate how to harness negative samples (i.e., examples where the model is prone to making errors) to aid in selecting positive sample examples for constructing more effective prompt demonstrations, with the goal of enhancing the LLM's context learning performance. We introduce an intuitive method: for each test question, we initially retrieve positive and negative examples from their respective corpora based on semantic similarity. Subsequently, we retrieve the most similar posi-tive example from the positive corpus based on the negative examples and concate-nate the two positive examples retrieved through these steps to form the final demon-stration.

Our motivation is to view the negative sample example corpus as a collection of incorrect answers. Given that these types of negative samples are likely to be an-swered incorrectly, we seek to practice these specific areas, i.e., mock questions. By combining the genuine example retrieved first with the positive example retrieved based on the negative sample secondly, we aim to not only practice questions that the model is already proficient at, but also focus on questions that are prone to errors.
We conducted experiments on seven datasets spanning three reasoning tasks. The results demonstrate that our proposed method significantly surpasses the comparison methods, attesting to the effectiveness of our approach. By utilizing negative sam-ples, we can construct superior prompt examples and enhance context learning per-formance. Additionally, we performed a quantitative experimental analysis on one selected dataset from each of the three reasoning tasks to determine the optimal number of negative samples to use. Our approach of leveraging negative samples marks progress in the development of more effective context examples.
We believe that this iterative approach of constructing retrieval corpora could serve as a versatile framework.

\bibliographystyle{splncs04}
\bibliography{ref}

\end{document}